\documentclass{article}
\PassOptionsToPackage{numbers, compress}{natbib}
\usepackage[preprint]{neurips_2025}
\usepackage[utf8]{inputenc} 
\usepackage[T1]{fontenc}    
\usepackage{hyperref}       
\usepackage{url}            
\usepackage{booktabs}       
\usepackage{amsfonts}       
\usepackage{nicefrac}       
\usepackage{microtype}      
\usepackage{xcolor}         
\usepackage{graphicx}
\usepackage{amsmath,amssymb,amsfonts}
\usepackage[capitalize,noabbrev,nameinlink]{cleveref}

\usepackage{algorithm}
\usepackage{algpseudocode}

\usepackage{bbm}
\newtheorem{hypothesis}{Hypothesis}
\newtheorem{definition}{Definition}
\usepackage{afterpage}

\usepackage[textsize=tiny]{todonotes}
\newcommand\scalemath[2]{\scalebox{#1}{\mbox{\ensuremath{\displaystyle #2}}}}

\title{SE-Merging: A \underline{S}elf-\underline{E}nhanced Approach for Dynamic Model Merging}

\author{%
Zijun Chen\textsuperscript{1}$^{\ast}$, Zhanpeng Zhou\textsuperscript{1}$^{\ast}$, Bo Zhang\textsuperscript{2}, Weinan Zhang\textsuperscript{1}, Xi Sun\textsuperscript{3\dag}, Junchi Yan\textsuperscript{1,4\dag}\\
\textsuperscript{1}School of Computer Science, Shanghai Jiao Tong University, \\
\textsuperscript{2}Shanghai Artificial Intelligence Laboratory, 
\textsuperscript{3}MetaLight HK Limited, \\
\textsuperscript{4}Shanghai-Chongqing Institute of Artificial Intelligence, Shanghai Jiao Tong University\\
  \texttt{\{chenzijun,zzp1012,yanjunchi\}@sjtu.edu.cn}\\
  $^{\ast}$ Equal Contribution, $\dag$ Correspondence Authors
}

\begin{document}

\maketitle

\begin{abstract}
  Model merging has gained increasing attention due to its intriguing property: interpolating the parameters of different task-specific fine-tuned models leads to multi-task abilities. However, despite its empirical success, the underlying mechanisms of model merging remain poorly understood. In this work, we delve into the mechanism behind model merging from a representation perspective. Our analysis reveals that model merging achieves multi-task abilities through two key capabilities: i) distinguishing samples from different tasks, and ii) adapting to the corresponding expert model for each sample. These two capabilities allow the merged model to retain task-specific expertise, enabling efficient multi-task adaptation. Building on these insights, we propose \texttt{SE-Merging}, a self-enhanced model merging framework that leverages these two characteristics to dynamically identify the corresponding task for each sample and then adaptively rescales the merging coefficients to further enhance task-specific expertise in the merged model. Notably, \texttt{SE-Merging} achieves dynamic model merging without additional training. Extensive experiments demonstrate that \texttt{SE-Merging} achieves significant performance improvements while remaining compatible with existing model merging techniques.
\end{abstract}

\section{Introduction}

Recent works on model merging have drawn considerable attention, particularly with the rise of Large Language Models~\citep{qwen2025qwen25technicalreport, brown2020languagemodelsfewshotlearners}. 
The core idea of model merging is to integrate the parameters of multiple task-specific expert models, each independently fine-tuned on different tasks, into a single unified model that can perform multiple tasks. 
Specifically, let $\mathcal{T}=\{T_i\}_{i=1}^T$ denote the set of different tasks and $\boldsymbol{\theta}_i$ denote the expert model fine-tuned over task $T_i$, the simplest form of model merging is given by \begin{equation}\label{eq:simple_avg}
    \boldsymbol{\theta}_{\rm Merged} = \sum_{i=1}^T \lambda_i \boldsymbol{\theta}_i, \quad\lambda_i \in [0, 1],
\end{equation}
where $\lambda_i$ denote the scaling coefficient associated with each expert model.
Model merging enables multi-task capabilities without additional training and significantly reduces inference costs compared to model ensembles~\cite{dong2020modelensemblesurvey}.
The growth of open-source platforms like Hugging Face~\citep{wolf2020huggingfacestransformersstateoftheartnatural} has also facilitated the development of model merging methods by providing easy access to a wide range of fine-tuned models for various tasks.

Despite the empirical success of the model merging methods, our understanding of them still lags behind.
Several attempts have been made to understand the mechanisms behind model merging; however, many such works focus on the linearization of neural networks.
Ortiz-Jimenez et al.~\cite{ortiz2024task} tried to explain the model merging in the lazy regime~\cite{Chizat2019lazytraining}, where the training dynamics are linear and can be captured by a static kernel function, commonly known as the neural tangent kernel (NTK)~\cite{jacot2018neural}.
Furthermore, Zhou et al.~\cite{zhou2024on} proposed the Cross-Task Linearity, where the model essentially functions as a linear map in the internal representation space.
Understanding model merging beyond linearization is of paramount importance and remains an active area of research.

\begin{figure*}[!tb]
  \begin{center}
    \includegraphics[width=0.91\textwidth]{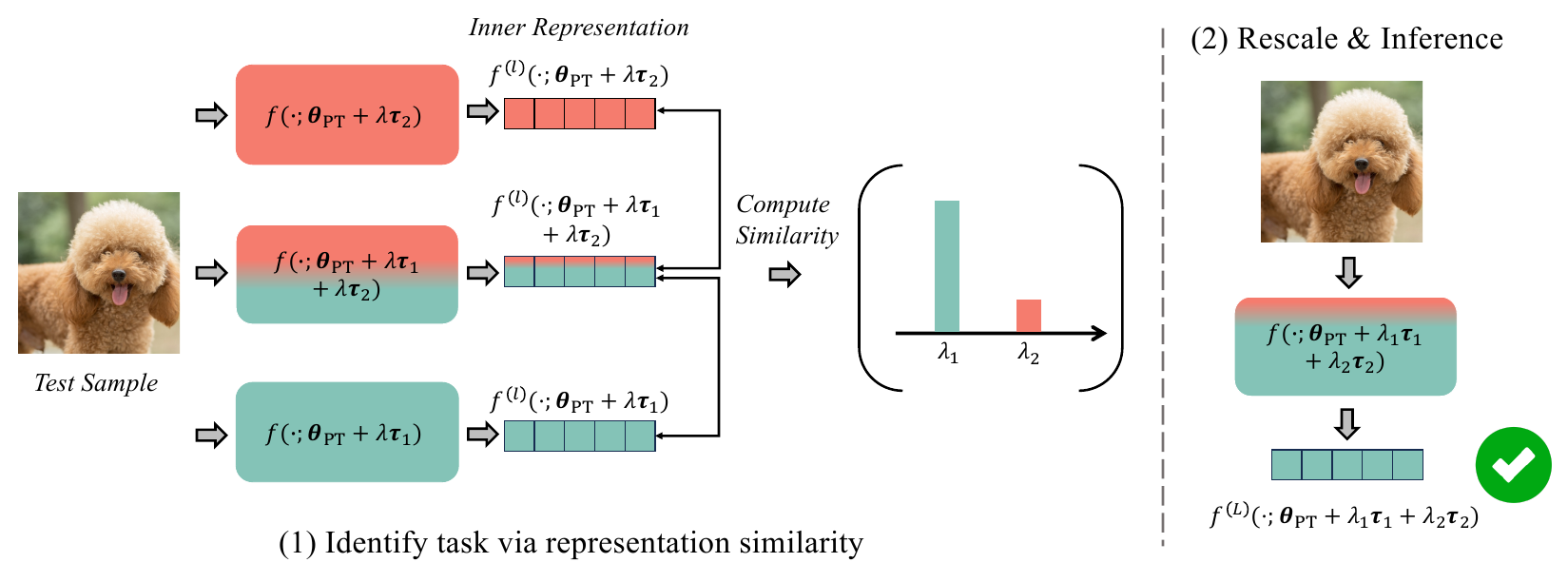}
    \caption{
    \textbf{Overview of our self-enhanced model merging framework.} 
    \textbf{(1) Left}: Following Task arithmetic~\cite{ilharco2023editing}, we first obtain the merged model via task addition, i.e. $\boldsymbol{\theta}_{\rm PT} + \lambda \boldsymbol{\tau}_1 + \lambda \boldsymbol{\tau}_2$, and the fine-tuned model via $\boldsymbol{\theta}_{\rm PT} + \lambda \boldsymbol{\tau}_1$ and $\boldsymbol{\theta}_{\rm PT}+ \lambda \boldsymbol{\tau}_2$. 
    Here, $\boldsymbol{\tau}_1=\boldsymbol{\theta}_1-\boldsymbol{\theta}_{\rm PT}$ and $\boldsymbol{\tau}_2 = \boldsymbol{\theta}_2-\boldsymbol{\theta}_{\rm PT}$ denote the task vectors of task $T_1$ and $T_2$ respectively. $\boldsymbol{\theta}_{\rm PT}$ denotes the pre-trained model and $\boldsymbol{\theta}_{\rm 1}$ and $\boldsymbol{\theta}_{\rm 2}$ denote the models fine-tuned on task $T_1$ and $T_2$ respectively.
    Then, for the sample $\boldsymbol{x}$ from an unknown task $T$, we compute the representation of merged model and that of fine-tuned models. 
    Based on our Representation Auto-Adaptation Hypothesis~\ref{hypo:hypo1}, the merged model implicitly adapts to the expert model for task $T$ and the representation of merged model is more similar to that of the fine-tuned model for task $T$.
    Thus, we can identify the actual task to which the sample $\boldsymbol{x}$ belongs by calculating the representation similarity, that is $T=T_1$.
    Next, we give the new scaling coefficient for sample $\boldsymbol{x}$, assigning larger coefficient to the \textit{task vector} of task $T_1$. 
    \textbf{(2) Right}: We merge the model specific for sample $\boldsymbol{x}$ using the new merging coefficient. 
    The final output $f^{(L)}(\boldsymbol{x}; \boldsymbol{\theta}_{\rm PT} + \lambda_1\boldsymbol{\tau}_1 + \lambda_2\boldsymbol{\tau}_2)$ resembles more to the $f^{(L)}(\boldsymbol{x}; \boldsymbol{\theta}_{\rm PT} + \lambda\boldsymbol{\tau}_1)$, retaining more task-specific information and thus further enhancing the performance. 
    Different colors (\textcolor{red}{red} and \textcolor{green}{green}) represent different tasks. 
    } 
     \label{fig:main_fig}
  \end{center}
  \vspace{-1.0em}
\end{figure*}

\textbf{Internal Representation Analysis.}
In this work, we take a representation perspective to investigate the underlying mechanism behind model merging.
Surprisingly, we identify two key capabilities in the merged model to enable multi-task abilities: 
\begin{itemize}
    \item \emph{Capability I.} Distinguishes samples from different tasks.
    \item \emph{Capability II.} Adapt to the corresponding expert model for each sample.
\end{itemize}
Denote $f^{(\ell)}(\boldsymbol{\theta})$ as the representation of model $\boldsymbol{\theta}$ at $\ell$-th layer.
For Capability I, we observe that the representations of the merged model $f^{(\ell)}(\boldsymbol{\theta}_{\rm Merged})$ form different clusters according to the task of each sample.
For Capability II, we find that, given an input sample $\boldsymbol{x}$ from task $T_i$, the representation of the merged model $f^{(\ell)}(\boldsymbol{\theta}_{\rm Merged})$ is closer to that of the expert model fine-tuned over task $T_i$, i.e., $f^{(\ell)}(\boldsymbol{\theta}_{i})$.
These two capabilities together give rise to the effectiveness of model merging, which we formalize into the \emph{Representation Auto-Adaptation Hypothesis} (see Hypothesis~\ref{hypo:hypo1}).
We follow the settings of Task Arithmetic method~\citep{ilharco2023editing} and validate our hypothesis through extensive experiments.
Furthermore, we take a first attempt to theoretically understand the Representation Auto-Adaptation Hypothesis by assuming the Weight Disentanglement~\cite{ortiz2024task}.

\textbf{From Discovery to Applications.}
Based on the Representation Auto-Adaptation Hypothesis, we propose \texttt{SE-Merging}, a self-enhanced model merging framework.
Specifically, by leveraging representation similarity, \texttt{SE-Merging} dynamically identifies the task that each sample belongs to and then adaptively adjusts the importance of each fine-tuned model during the merging process.

See \cref{fig:main_fig} for an overview of our method.
\texttt{SE-Merging} requires no additional training and can be seamlessly integrated into existing pipelines, such as Task Arithmetic~\citep{ilharco2023editing}.
Extensive experiments on both vision and language tasks validate the effectiveness of \texttt{SE-Merging}.

\textbf{Summary.}
Our contributions can be summarized as follows: 
\begin{itemize}
    \item We unveils the hidden mechanism of model merging and propose the \textit{Representation Auto-Adaptation Hypothesis}, along with  empirical evidence and theoretical analysis.
    \item We propose \texttt{SE-Merging}, a self-enhanced training-free model merging framework based on the \textit{Representation Auto-Adaptation Hypothesis}, showing significant performance gain across vision and language tasks.
\end{itemize}

\begin{figure*}[!tb]
  \begin{center}
    \includegraphics[width=\textwidth]{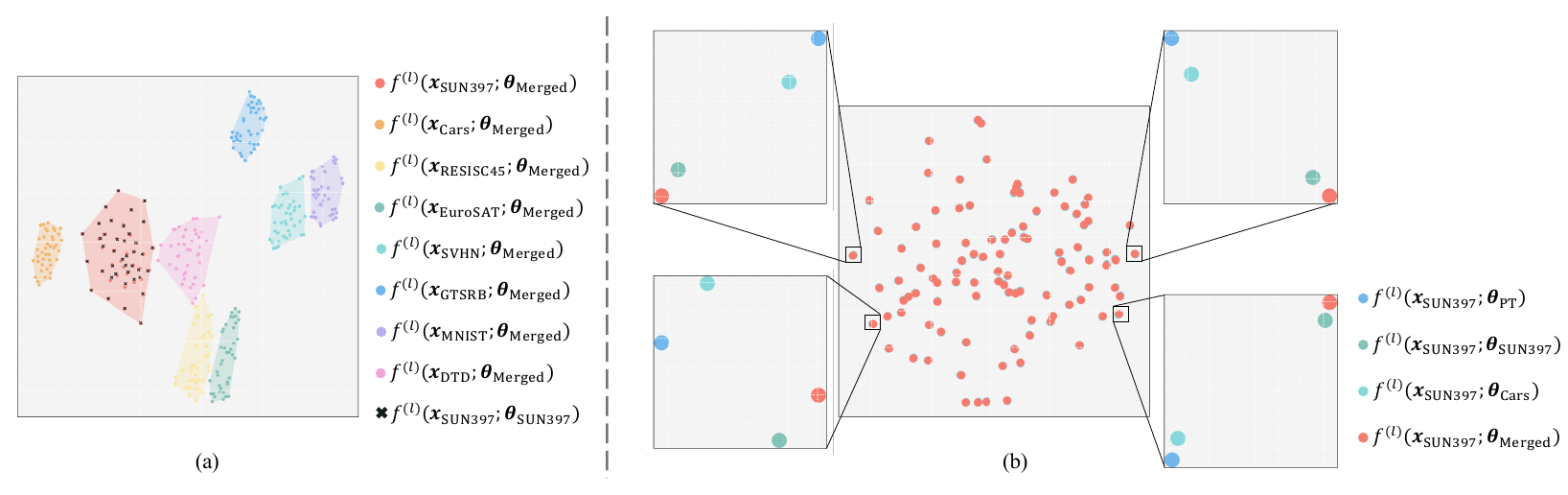}
    \caption{
    \textbf{t-SNE visualization illustrating the two key capabilities}.
    \textbf{(a) Capability I: Distinguish samples from different tasks.}
    The representations of the merged model on eight vision tasks, i.e. $f^{(\ell)}(\boldsymbol{x}; \boldsymbol{\theta}_{\rm Merged}), {\rm where\ } \boldsymbol{x} \in \cup_{j\in [T]}\mathcal{D}_j$, are seperated and located in different clusters.
    In addition, for SUN397 dataset, the representations $f^{(\ell)}(\boldsymbol{x_{\text{SUN397}}}; \boldsymbol{\theta_{\text{SUN397}}})$ and $f^{(\ell)}(\boldsymbol{x_{\text{SUN397}}}; \boldsymbol{\theta_{\text{Merged}}})$ are located in the same cluster.
    These results imply that the merged model can distinguish samples from different tasks.
    \textbf{(b) Capability II: Adapt to the corresponding expert model for each sample.}
    we merge two models $\boldsymbol{\theta_{\text{SUN397}}}$ and $\boldsymbol{\theta_{\text{Cars}}}$. 
    For the sample from SUN397 dataset, $f^{(\ell)}(\boldsymbol{x_{\text{SUN397}}}; \boldsymbol{\theta_{\text{SUN397}}})$ is closer to $f^{(\ell)}(\boldsymbol{x_{\text{SUN397}}}; \boldsymbol{\theta_{\text{Merged}}})$, while $f^{(\ell)}(\boldsymbol{x_{\text{SUN397}}}; \boldsymbol{\theta_{\text{Cars}}})$ is closer to $f^{(\ell)}(\boldsymbol{x_{\text{SUN397}}}; \boldsymbol{\theta_{\text{PT}}})$. 
    This demonstrates that the merged model shares a similar inference process as the corresponding expert model.
    Here, $\boldsymbol{\theta_{\text{SUN397}}}$ and $\boldsymbol{\theta_{\text{Cars}}}$ denote the model fine-tuned on $\text{SUN397}$ and $\text{Cars}$ from the pre-trained model $\boldsymbol{\theta_{\text{PT}}}$.  $\boldsymbol{x_{\text{SUN397}}}$ and $\boldsymbol{x_{\text{Cars}}}$ denote the sample from dataset $\text{SUN397}$ and $\text{Cars}$.
    }
     \label{fig:fig2}
  \end{center}
  \vspace{-1.0em}
\end{figure*}

\subsection{Related Work}

\textbf{Model Merging Methods.}
Recently, many model merging algorithms have been proposed.
Model Soups~\cite{wortsman2022modelsoups} considered the simplest form of model merging in \cref{eq:simple_avg}. 
Task Arithmetic~\cite{ilharco2023editing} introduced the concept of a \textit{task vector}, defined as the difference in weights between a fine-tuned model and the pre-trained model, and incorporates these \textit{task vectors} into the pre-trained model via simple arithmetic.
Despite the effectiveness of these two methods, two challenges arise from two key aspects: i) determining the appropriate scaling coefficients for each expert model, i.e., $\{\lambda_i\}_{i=1}^T$  and ii) resolving parameter conflicts when combining the weights of different models.

To determine the scaling coefficients, Fisher-Merging~\cite{matena2022fishermerging} leveraged the Fisher information matrix to evaluate the importance of individual parameters and used this importance as a coefficient to weight the parameters during the merging process.
RegMean~\cite{jin2023regmean} proposed a closed-form solution for the merging coefficients by solving a linear regression problem for the linear layers in the model.
AdaMerging~\cite{yang2024adamerging} employed gradient descent optimization to learn merging coefficients by minimizing entropy as a surrogate loss on unlabeled test data.
However, these methods demand significant computational resources and require access to the downstream task data.

To address the parameter conflicts problem, TIES-Merging ~\citep{yadav2024ties} proposed to preserve the large-magnitude parameters and resolve sign conflicts in the merging weights. 
DARE~\citep{yu2024dare} sparsified the model's parameters and subsequently performs rescaling on the sparse parameters.
PCB-Merging~\citep{du2024parameter} attempted to alleviate performance drops by analyzing intra-balancing of parameters within a single parameter and inter-balancing across parameters.

\textbf{Theoretical Advances on Model Merging.} 
Many existing works have tried to understand model merging theoretically. 
Ortiz-Jimenez et al.~\cite{ortiz2024task} investigate the merging of fine-tuned models on different datasets using the neural tangent kernel (NTK)~\cite{jacot2018neural}, highlighting weight disentanglement as a necessary condition for effective model merging.
Furthermore, Zhou et al.~\cite{zhou2024on} provide an explanation through Cross-Task Linearity (CTL), demonstrating that the representations in the merged model are approximately equivalent to the linear interpolation of representations from the two fine-tuned models at each layer.

Additionally, model merging is closely related to the loss landscape of deep neural networks. 
Prior studies~\cite{draxler2018essentially,garipov2018loss,frankle2020linear,entezari2021role} find that different minima in the loss landscape can be connected via a (non-)linear path of nearly constant loss, often referred to as \emph{(Linear) Model Connectivity}.
Several papers~\cite{freeman2017topology,liang2018understanding,venturi2018spurious,nguyen2018loss,nguyen2019connected,kuditipudi2019explaining, zhou2023going} also tried to prove the (Linear) Mode Connectivity under various settings.

\textbf{Dynamic Model Merging Methods.} 
Recently, researchers have developed dynamic merging methods that outperform traditional static model merging approaches. 
EMR-Merging~\cite{huang2024emrmerging} separated model merging into two parts: a shared component for all tasks and task-specific components. 
It uses a ``perfect router'' to choose the right task-specific module for each input, reducing conflicts between tasks.
Twin-Merging~\cite{lu2024twinmerging} used a similar approach, adaptively combining shared and task-specific modules during inference based on routing scores.
Weight-Ensembling MoE~\cite{tang2024merging} is inspired by the Mixture-of-Experts (MoE) framework~\cite{shazeer2017moe}. 
It adds trainable routers at each layer of the model, allowing it to dynamically merge weights layer by layer.
Representation Surgery~\cite{yang2024representation} introduces a ``surgery module'' for each task, which adjusts the merged model's outputs to better match the outputs of the fine-tuned model for that specific task. 
These methods aim to make model merging more flexible and effective.

\section{Preliminaries} \label{sec:preliminary}
\textbf{Notations.} 
Denote $[k] = \{1, 2, \cdots, k\}$.
We consider an $L$-layer neural network of the form $\hat{y} = f(\boldsymbol{x}; \boldsymbol{\theta}) \in \mathbb{R}^c$, with $\boldsymbol{\theta} \in \mathbb{R}^p$ representing the model parameters and $\boldsymbol{x} \in \mathbb{R}^d$ representing the input. 
Here, $p$ is the number of parameters, $d$ is the input dimension, and $c$ is the output dimension, which corresponds to the number of classes in classification tasks.
$f^{(\ell)}( \boldsymbol{x}; \boldsymbol{\theta}) \in \mathbb{R}^{d_\ell}$ represents the internal representation of the neural network at the $\ell$-th layer for input $\boldsymbol{x}$. 
Here, $d_\ell$ denotes the dimension of the $\ell$-th layer ($0\le \ell \le L$) and $f^{(L)}(\boldsymbol{x}; \boldsymbol{\theta}) = f(\boldsymbol{x}; \boldsymbol{\theta})$.
When the context is clear, we also use $f^{(\ell)}(\boldsymbol{\theta})$ as a shorthand for $f^{(\ell)}(\cdot; \boldsymbol{\theta})$.

\textbf{Problem Setup.} 
We formalize our problem setup of model merging as Task Arithmetic~\cite{ilharco2023editing}. 
Consider a pre-trained model $\boldsymbol{\theta}_{\rm PT}$ and a collection of different tasks $\mathcal{T} =\{T_i\}_{i=1}^T$, where each task $T_i$ is associated with a dataset $\mathcal{D}_i = \{(\boldsymbol{x}_k^i, y_k^i)\}_{k=1}^{N_i}$.
For each task $T_i \in \mathcal{T}$, we fine-tune the pre-trained model to obtain a task-specific expert model, denoted as $\boldsymbol{\theta}_i$.
The central idea of model merging is to combine the set of fine-tuned models $\{\boldsymbol{\theta}_i\}_{i=1}^{T}$ into a single unified model, parameterized as $\boldsymbol{\theta}_{\rm{Merged}}$, which is capable of handling all tasks effectively.

We follow the convention in Task Arithmetic~\cite{ilharco2023editing}, formulating the model merging using \emph{task vectors}. 
For each task $T_i \in \mathcal{T}$, the task vector is defined as $\boldsymbol{\tau}_i = \boldsymbol{\theta}_i - \boldsymbol{\theta}_{\rm{PT}}$. 
Then the parameterization of the merged model can be written as:
$$\boldsymbol{\theta}_{\rm{Merged}} = \boldsymbol{\theta}_{\rm{PT}} + \sum_{i}^{T} \lambda_i \boldsymbol{\tau}_i$$
where $\lambda_i$ is the scaling coefficient for each task vector $\boldsymbol{\tau}_i$. 

Without loss of generality, we also refer to \emph{fine-tuned model} as the sum of a pre-trained model and a \textit{task vector} scaled by a coefficient, i.e.,
$$\boldsymbol{\theta}_{\rm{PT}} + \lambda \boldsymbol{\tau}, \lambda > 0$$ where $\boldsymbol{\theta}_{\rm PT}$ represents the pre-trained model, $\boldsymbol{\tau}$ is the \textit{task vector}, and $\lambda$ is a scaling coefficient.

\textbf{Main Experimental Setup.} 
\label{exp_setup}
We conduct our experiments on both the vision and language tasks. 
For vision tasks, we adopt the experimental setup in Task Arithmetic \cite{ilharco2023editing}. 
We employ ViT-B/32 and ViT-L/14 vision encoders in CLIP \cite{radford2021learning} as pre-trained models, and then fine-tune them on eight image classification datasets, respectively: SUN397~\cite{SUN397}, Cars~\cite{Cars}, RESISC45~\cite{RESISC45}, EuroSAT~\cite{EuroSAT}, SVHN~\cite{SVHN}, GTSRB~\cite{GTSRB}, MNIST~\cite{MNIST} and DTD~\cite{DTD}. 
For language tasks, we follow the setup in FusionBench \cite{tang2024fusionbenchcomprehensivebenchmarkdeep}. 
We choose the GPT-2 \cite{gpt2} as the pre-trained model and fine-tune the model on seven language tasks from the GLUE~\cite{wang-etal-2018-glue} benchmark: CoLA~\cite{warstadt-etal-2019-cola},  STS-2~\cite{socher-etal-2013-sst2}, MRPC~\cite{dolan-brockett-2005-mrpc}, QQP~\cite{shankar2017qqp}, MNLI~\cite{williams-etal-2018-mnli}, QNLI~\cite{rajpurkar-etal-2016-qnli} and RTE~\cite{giampiccolo-etal-2007-rte}. 

\section{Towards Understanding of Model Merging from a Representation Perspective.}\label{sec:understanding}

In this section, we present a representation analysis of model merging. 
Specifically, we identify two key capabilities needed to perform model merging. 
Extensive experimental evidence and theoretical understandings support our findings.

\textbf{Capability I: Distinguish samples from different tasks.}
In \cref{fig:fig2}~(a), we visualize the representations of the merged model over samples from different tasks using the t-SNE algorithm~\cite{tsne}.
We observe that the representations of the merged model form different clusters based on the task of each sample.
This result indicates that the merged model could categorize the samples by tasks without explicitly acquiring the task information.
Furthermore, we find that for a specific task $T_i$, the representations of the merged model reside in the same cluster as those of the model fine-tuned over task $T_i$.
This result motivates our discovery of the Capability II.

\textbf{Capability II: Adapt to the corresponding expert model for each sample.}
To further investigate the representations of the merged model, we consider a simplified case where only two fine-tuned models are merged.
Without loss of generality, we choose models fine-tuned over the SUN397 dataset and the Cars dataset.
In \cref{fig:fig2}~(b), we visualize the representations of the finetuned models and the merged model.
In addition, we also include the representations of the pre-trained model for comparison.
For each sample $\boldsymbol{x}$ of task $T_i$, we show that the representation of the merged models is closer to that of the model fine-tuned over task $T_i$. 
This result implies that given a sample $\boldsymbol{x}$ of task $T_i$, the merged model shares a similar inference process as the model fine-tuned over task $T_i$, suggesting its ability to automatically adapt to the corresponding expert model for the input sample.

\textbf{Representation Auto-Adaptation Hypothesis.}
Based on the two key capabilities, we propose the following Hypothesis~\ref{hypo:hypo1}, which formalizes our main discovery.
\begin{hypothesis}[Representation Auto-Adaptation Hypothesis]\label{hypo:hypo1}
For a given sample $\boldsymbol{x}_i$ of task $T_i$, the representation of the merged model at $\ell$-th layer closely resembles that of the model fine-tuned over task $T_i$:
\begin{equation}
    \rm{Dist}(f^{(\ell)}(\boldsymbol{x}_i; \boldsymbol{\theta}_{\rm{Merged}}), f^{(\ell)}(\boldsymbol{x}_i; \boldsymbol{\theta}_{\rm{PT}} + \lambda \boldsymbol{\tau}_i)) \approx 0.
\end{equation}
where $\boldsymbol{\theta}_{\rm{Merged}} = \boldsymbol{\theta}_{\rm{PT}} + \lambda \sum_{i}^{T} \boldsymbol{\tau}_i$, $\rm{Dist}(\cdot, \cdot)$ is the distance measure, e.g., $\ell_2$ norm or cosine distance.
Informally, the hypothesis can be expressed as $f^{(\ell)}(\boldsymbol{x}_i; \boldsymbol{\theta}_{\rm{Merged}}) \approx f^{(\ell)}(\boldsymbol{x}_i; \boldsymbol{\theta}_{\rm{PT}} + \lambda \boldsymbol{\tau}_i)$, meaning that these two representations are similar in the representation space.
\end{hypothesis}

\textbf{More empirical evidence: From visualization to quantitative study.} 
Extending from the visualization in \cref{fig:fig2}, we conduct a quantitative study to further validate our Hypothesis~\ref{hypo:hypo1}. 
Specifically, we measure the precision with which the merged model generates representations that closely align with those of the corresponding expert model, as opposed to those of models fine-tuned on unrelated tasks.

\textbullet~\emph{Experimental setup.} 
Following the standard setting in Task Arithmetic \cite{ilharco2023editing}, we conduct experiments on both ViT-B/32 and ViT-L/14, with the eight image classification tasks mentioned in ~\cref{sec:preliminary}. 
The merged model is computed by task addition~\cite{ilharco2023editing}, i.e., $\boldsymbol{\theta_{\text{Merged}}} = \boldsymbol{\theta_{\text{PT}}} + \lambda \sum_{i=1}^{T} \boldsymbol{\tau}_i$, with $\lambda=0.3$ for each task. 

\begin{figure*}[!tb]
  \begin{center}
    \includegraphics[width=1.0\textwidth]{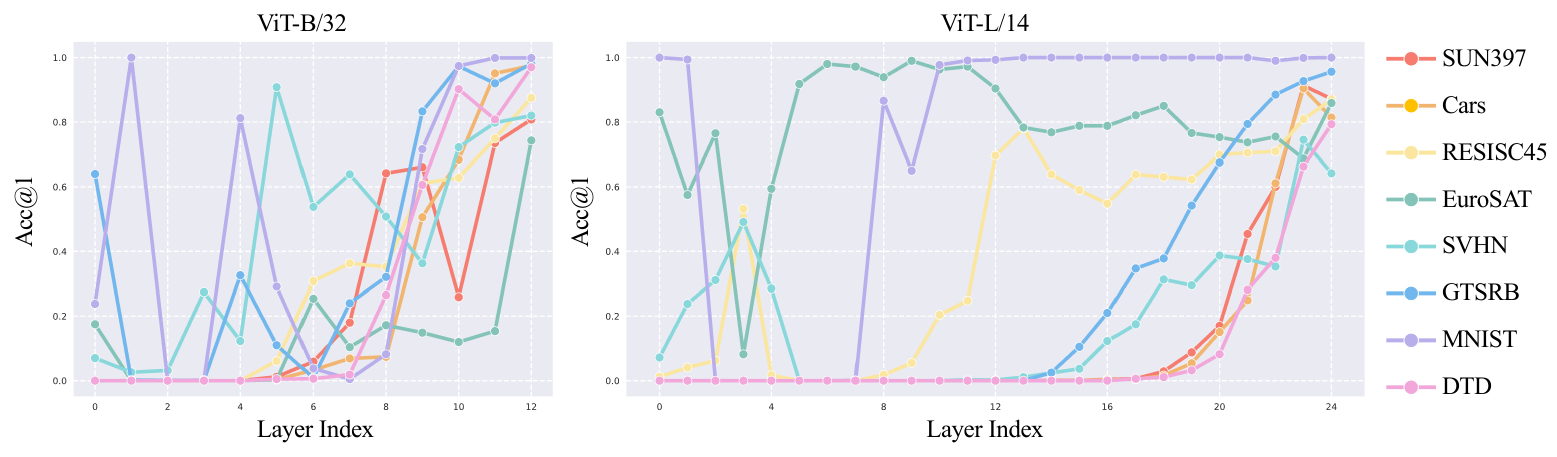}
    \caption{
    \textbf{Quantitative validation of Hypothesis~\ref{hypo:hypo1}.} 
    We report the \texttt{acc@1} for both ViT-B/32 and ViT-L/14 across all tasks and different layers. 
    The \texttt{acc@1} metric is computed as described in \cref{alg:pilot}.
    As shown, at deeper layers, e.g., $12$-th layer for ViT-B/32 and $24$-th layer for ViT-L/14, \texttt{acc@1} approaches 1.0 for each task, indicating that for most of the samples in the task, the representation of merged model is closely aligned with that of the corresponding fine-tuned model. 
    Moreover, this result demonstrates that for most of the tasks, the task-specific representation is obtained at deeper layers. 
    }
     \label{fig:fig_pilot}
  \end{center}
  \vspace{-1.0em}
\end{figure*}

\textbullet~\emph{Experimental details.}
Given a sample $\boldsymbol{x}$ of task $T_i$, we compute the representation of the merged model at $\ell$-th layer, i.e., $f^{(\ell)}(\boldsymbol{x}; \boldsymbol{\theta}_{\rm{Merged}})$.
We also obtain the representations of the fine-tuned models, i.e., $f^{(\ell)}(\boldsymbol{x}; \boldsymbol{\theta}_{\rm PT} + \lambda \boldsymbol{\tau}_j), j\in[T]$. 
Then, we measure the distance between the representation of the merged model and those of fine-tuned models, i.e., $\Vert f^{(\ell)}(\boldsymbol{x}; \boldsymbol{\theta}_{\rm{Merged}}) - f^{(\ell)}(\boldsymbol{x}; \boldsymbol{\theta}_{\rm PT} + \lambda \boldsymbol{\tau}_j) \Vert_2, j \in [T]$.
We sort the distances in ascending order and determine the rank $rk$ of $\Vert f^{(\ell)}(\boldsymbol{x}; \boldsymbol{\theta}_{\rm{Merged}}) - f^{(\ell)}(\boldsymbol{x}; \boldsymbol{\theta}_{\rm PT} + \lambda \boldsymbol{\tau}_i) \Vert_2$ within the sorted distance sequence. 
A lower rank indicates that the representation of the merged model for input $\boldsymbol{x}$ is closer to that of the model fine-tuned over task $T_i$.
For all samples in task $T_i$, we use \texttt{acc@k} as an evaluation metric to measure how well the merged model adapts to the expert model for that task.
\texttt{acc@k} is defined as the proportion of samples in task $T_i$ whose rank $rk$ is smaller than \texttt{k}.
The high \texttt{acc@k} indicates that, for the vast majority of samples in task $T_i$, the representation of the merged model is most similar to that of the corresponding fine-tuned model.
See~\cref{alg:pilot} for more details.

\textbullet~\emph{Experimental analysis.}
In ~\cref{fig:fig_pilot}, the \texttt{acc@1} scores are consistently close to 1.0 for both ViT-B/32 and ViT-L/14 across all tasks at deep layers. 
This result strongly supports our Hypothesis~\ref{hypo:hypo1} that for each task $T_i$, the merged model effectively resembles the corresponding fine-tuned model $\boldsymbol{\theta}_i$ in terms of the inner representations.
Additionally, for most tasks, \texttt{acc@1} in shallow layers is lower than in deeper layers, suggesting that earlier layers capture more general, shared representations, while deeper layers retain task-specific representations.

\begin{algorithm}[!tb]
\small
    \caption{Empirical Study: Validating the Representation Auto-Adaptation Hypothesis}
    \begin{algorithmic}[1]
    \Require
    \quad Model $f(\cdot; \boldsymbol{\theta})$, pre-trained weights $\boldsymbol{\theta}_{\rm{PT}}$, 
    $T$ task vectors $\{\boldsymbol{\tau}_t\}_{t=1}^{T}$, 
    Dataset $\mathcal{D}_i = \{(\boldsymbol{x}_n^{i}, y_n^i)\}_{n=1}^{N}$ with $N$ samples for task $T_i$,
    the layer $\ell$ of the model.
    
    \Ensure
    \quad \texttt{acc@k} metric for evaluating the adaptation of merged model to the expert model for task $T_i$.

    \State {Initialize merged model with weights: 
    \State \quad $\boldsymbol{\theta}_{\rm{Merged}} = \boldsymbol{\theta}_{\rm{PT}} + \lambda \sum_{i=1}^{T} \boldsymbol{\tau}_i$, where $\lambda=0.3$.}
    
    \For{$n$-th sample $\boldsymbol{x}^i_n \in \mathcal{D}_i$}
        \State {Compute representations for sample $\boldsymbol{x}^i_n$ on merged model and fine-tnued models:}
        \State \quad $r_{\rm{Merged}} = f^{(\ell)}(\boldsymbol{x}^i_n; \boldsymbol{\theta}_{\rm{Merged}})$
        \State \quad $r_t = f^{(\ell)}(\boldsymbol{x}^i_n; \boldsymbol{\theta}_{\rm PT} + \lambda \boldsymbol{\tau}_t)$, $t \in [T]$
        
        \State Compute the $\ell_2$ distances of representations between merged model and fine-tuned models:
        \State \quad $d_t = \|r_{\rm{Merged}} - r_t\|_2$, $t \in [T]$
        
        \State Sort distances $\{d_t\}_{t=1}^{T}$ in ascending order and determine rank of $d_i = \|r_{\rm{Merged}} - r_i\|_2$ within the ordered distance sequence for sample $\boldsymbol{x}^i_n$:
        \State \quad   $rk_n = \text{rank}(d_{i})$ in $\text{AscOrdered}(\{d_t\}_{t=1}^{T})$
    \EndFor
    
    \State Compute \texttt{acc@k}:
    \State \quad $\texttt{acc@k} = \frac{1}{N} \sum_{n=1}^{N} \mathbbm{1}[rk_n \leq \texttt{k}]$
    
    \State \Return \texttt{acc@k} metric.
    \end{algorithmic}
    \label{alg:pilot}
\end{algorithm}

\textbf{Theoretical understandings: Insights into Hypothesis~\ref{hypo:hypo1}.}
We also take an initial attempt to theoretically understand our discovery.
We first recall the definition of Weight Disentanglement in \cref{def:weight_disent}.

\begin{definition}[Weight Disentanglement~\cite{ortiz2024task}]\label{def:weight_disent}
    A model $f$ is weight disentangled with respect to a set of task vectors $\{\boldsymbol{\tau}_i\}_{i=1}^T$ and the corresponding supports $\{\mathcal{D}_i\}_{i=1}^T$  if \begin{align}
        f\left(\boldsymbol{x}; \boldsymbol{\theta}_{\rm PT} + \sum_{i=1}^T \alpha_i \boldsymbol{\tau}_i \right) = \sum_{i=1}^T g_i (\boldsymbol{x}; \alpha_i \boldsymbol{\tau}_i) + g_0 (\boldsymbol{x}).
    \end{align}
    where $g_i (\boldsymbol{x}; \alpha_i \boldsymbol{\tau}_i) = 0$ for $\boldsymbol{x} \notin \mathcal{D}_i$, and $g_0(\boldsymbol{x})=0$ for $\boldsymbol{x} \in \cup_{i \in [T]} \mathcal{D}_i$.
\end{definition}

Ortiz-Jimenez et al.~\cite{ortiz2024task} argued that the weight disentanglement is the necessary condition to achieve model merging.
A trivial way to satisfy \cref{def:weight_disent} is to enforce the following decomposition of model $f$: \begin{equation}\label{eq:model_decomp}
    \begin{aligned}
        \scalemath{0.90}{f\left(\boldsymbol{x}; \boldsymbol{\theta}_{\rm PT} + \sum_{i=1}^T \alpha_i \boldsymbol{\tau}_i \right) = \sum_{i=1}^T f(\boldsymbol{x}; \boldsymbol{\theta}_{\rm PT} + \alpha_i \boldsymbol{\tau}_i) \mathbbm{1}(\boldsymbol{x} \in \mathcal{D}_i)}\\ + f(\boldsymbol{x}; \boldsymbol{\theta}_{\rm PT}) \mathbbm{1}(\boldsymbol{x} \notin \cup_{i \in [T]}\mathcal{D}_i). 
    \end{aligned}
\end{equation}
\cref{eq:model_decomp} indeed coincides with our Hypothesis~\ref{hypo:hypo1}, where for a sample $\boldsymbol{x} \in \mathcal{D}_i$, the model automatically adapts to the corresponding finetuned model $f(\boldsymbol{x}; \boldsymbol{\theta}_{\rm PT} + \alpha_i \boldsymbol{\tau}_i)$.
However, \cref{eq:model_decomp} only concerns the final output of the models, while our findings also consider their inner representations.

\section{\underline{S}elf-\underline{E}nhanced Model Merging: a Training-Free Sample-wise Method}

In this section, we introduce \underline{S}elf-\underline{E}nhanced Model Merging (\texttt{SE-Merging}), a training-free sample-wise method designed to improve model merging performance at inference time without requiring additional tuning.

\textbf{Methodology.} 
Our approach is based on Hypothesis~\ref{hypo:hypo1}, which states that the merged model generates representations closer to those of the corresponding fine-tuned model.
Instead of statically interpolating weights of fine-tuned models, we adopt a dynamic approach in the inference process for each sample.
Leveraging the Hypothesis~\ref{hypo:hypo1}, \texttt{SE-Merging} identifies the task to which the test sample belongs and assigning more importance to the corresponding \textit{task vector} $\boldsymbol{\tau}$ through rescaling the merging coefficients.
This process ensures that the rescaled merged model's parameter $\boldsymbol{\theta}_{\rm{SE-Merging}}$ aligned more closely with the task-specific fine-tuned model $\boldsymbol{\theta}_i$ in parameter space. 
Moreover, our approach does not rely on any routers~\cite{lu2024twinmerging, tang2024merging} that trained on test dataset to guide the merging process. 
Routers assist in the merging process by recognizing the task to which a test sample belongs and assigning greater importance to the corresponding \textit{task vector}, thereby improving merging performance.
\texttt{SE-Merging} inherently acts as an implicit router by leveraging inner representation similarity to identify the relevant task.
Furthermore, our approach is compatible with existing methods that address parameter conflicts, e.g., Task Arithmetic~\cite{ilharco2023editing}, TIES-Merging~\cite{yadav2024ties}.

\begin{algorithm}[!tb]
\small
    \caption{Self-Enhanced Model Merging}
    \label{alg:se_merging}
    \begin{algorithmic}[1]
        \Require Pre-trained model $\boldsymbol{\theta}_{\rm{PT}}$, $T$ task vectors of fine-tuned models $\{\boldsymbol{\tau}_t\}_{t=1}^{T}$, test set $\mathcal{D} = \{(\boldsymbol{x}_i, y_i)\}_{i=1}^{N}$, predefined scaling coefficient $\lambda$ and layer $\ell$. 
        
        \State Initialize merged model:
        \State \quad $\boldsymbol{\theta}_{\rm{Merged}} = \boldsymbol{\theta}_{\rm{PT}} + \lambda \sum_{t=1}^{T} \boldsymbol{\tau}_t$
        
        \For{each sample $\boldsymbol{x}_i \in \mathcal{D}$}
            \State  Compute representation of merged model and that of fine-tuned models:
            \State \quad 
            $r_{\rm{Merged}} = f^{(\ell)}(\boldsymbol{x}_i; \boldsymbol{\theta}_{\rm{Merged}})$
            \State \quad 
            $r_t = f^{(\ell)}(\boldsymbol{x}_i; \boldsymbol{\theta}_{\rm{PT}}+\lambda \boldsymbol{\tau}_t), \quad t \in [T]$
            \State {Compute $\ell_2$ distance:}
            \State \quad
            $d_t = \| r_{\rm{Merged}} - r_t \|_2, \quad t \in [T]$
            
            \State Convert distance to similarity \Comment{Reverse $\ell_2$ distance and perform min-max normalization}
            \State \quad
            $s_t = d_{\max} - d_t + d_{\min}, \quad t \in [T]$
            \State \quad
            $s_t^{\rm{norm}} = \frac{s_t - \min(s)}{\max(s) - \min(s)}, \quad t \in [T]$
            
            \State {Compute the rescaled merging coefficients:}
            \State \quad
            $\lambda_t = \frac{\exp(s_t^{\rm{norm}})}{\sum_{j=1}^{T} \exp(s_j^{\rm{norm}})}  T  \lambda, \quad t \in [T]$
            
            \State {Update the merged model for sample $\boldsymbol{x}_i$:}
            \State \quad
            $\boldsymbol{\theta}_{\rm{SE-Merge}} = \boldsymbol{\theta}_{\rm{PT}} + \sum_{t=1}^{T} \lambda_t \boldsymbol{\tau}_t$
            \State Evaluate sample $\boldsymbol{x}_i$ with merged model $\boldsymbol{\theta}_{\rm{SE-Merge}}$.
            
        \EndFor
    \end{algorithmic}
\end{algorithm}

As shown in the overview of~\cref{fig:main_fig}, the specific process of \texttt{SE-Merging} proceeds as follows:
For each sample $\boldsymbol{x}_i$ in $\mathcal{D}$, during inference stage, we perform four main steps. (1) Representation Computation. We compute the representation of the merged model and that of the fine-tuned models at $\ell$-th layer. 
(2) Representation Similarity Calculation. We use $\ell_2$ distance as a metric and compute the $\ell_2$ distance between the representation of the merged model and that of the fine-tuned models. Next, we transform the distance into similarity. 
(3) Merging Coefficients Rescaling. Based on the similarity scores, we dynamically rescale the merging coefficients, assigning larger weights to the relevant \textit{task vectors}. 
(4) Merging and Inference. We update the merged model for sample $\boldsymbol{x}_i$ and perform inference.
See Algorithm \ref{alg:se_merging} for more details.

\begin{table*}[!tb]
\centering
\caption{Multi-task performance when merging ViT-B/32 models on eight tasks. }
\label{tab:performance_vitb32} 
\resizebox{1.0\textwidth}{!}{
\begin{tabular}{l|cccccccc|c}
\toprule

{Methods}& SUN397 & Cars &RESISC45 & EuroSAT & SVHN & GTSRB & MNIST & DTD  & \textbf{Avg.} \\
\midrule
{Fine-tuned}  & 75.3  &  77.7  &  96.1  &  99.7  &  97.5  &  98.7  &  99.7  &  79.4 & 90.5   \\
{Multi-Task Learning} &    73.9  &  74.4  &  93.9  & 98.2    &  95.8  &  98.9   &  99.5   & 77.9 & 88.9  \\
\midrule
{Weight Averaging}~\cite{wortsman2022modelsoups} &  65.3  &  63.4  &  71.4  &  71.7  &  64.2  &  52.8  &  87.5  &  50.1  & 65.8 \\
{Fisher Merging}~\cite{matena2022fishermerging}&  68.6  &  69.2  &  70.7  &  66.4  &  72.9  &  51.1  &  87.9  &  59.9 & 68.3  \\
{RegMean}~\cite{jin2023regmean}  
&  65.3  &  63.5  &  75.6  &  78.6  &  78.1  &  67.4  &  93.7  &  52.0 & 71.8 \\
Task Arithmetic~\cite{ilharco2023editing}  
&63.8 & 62.1 & 72.0 & 77.6 & 74.4 & 65.1 & 94.0 & 52.2 & 70.1 \\
{Ties-Merging}~\cite{yadav2024ties}   &  64.8  &  62.9  &  74.3  &  78.9  &  83.1  &  71.4  &  97.6  &  56.2 & 73.6  \\
PCB-Merging~\cite{du2024parameter} & 65.5 & 64.1 & 78.1 & 80.2 & 84.7 & 77.1 & 98.0 & 58.4 & 75.8 \\
AdaMerging++~\cite{yang2024adamerging}   &66.6 &68.3 &82.2 &94.2 &89.6 &89.0 &98.3  &60.6 &81.1 \\
\midrule
\texttt{SE-Merging} (\textbf{Ours})
& \textbf{70.98} &  \textbf{71.0} & \textbf{88.03} & \textbf{94.44}  & \textbf{90.01}  & \textbf{96.13} & \textbf{99.04} & \textbf{69.95} & \textbf{84.96}  \\
\bottomrule
\end{tabular}
}
\end{table*}
\begin{table*}[!tb]
\centering
\caption{Multi-task performance when merging ViT-L/14 models on eight tasks.}
\label{tab:performance_vitl14} 
\resizebox{1.0\textwidth}{!}{
\begin{tabular}{l|cccccccc|c}
\toprule
{Methods}& SUN397 & Cars &RESISC45 & EuroSAT & SVHN & GTSRB & MNIST & DTD  & \textbf{Avg.} 

\\
\midrule
{Fine-tuned}  &  82.3  &  92.4  &  97.4  &  100  &  98.1  &  99.2  &  99.7  &  84.1  & 94.2   \\
{Multi-Task Learning}&80.8   &  90.6   &   96.3  & 96.3   & 97.6   & 99.1   &  99.6  &  84.4   & 93.5    \\
\midrule
{Weight Averaging}~\cite{wortsman2022modelsoups} &  72.1  &  81.6  &  82.6  &  91.9  &  78.2  &  70.7  &  97.1  &  62.8 & 79.6 \\
{Fisher Merging}~\cite{matena2022fishermerging}   &  69.2  &  88.6  &  87.5  &  93.5  &  80.6  &  74.8  &  93.3  &  70.0  & 82.2 \\
{RegMean}~\cite{jin2023regmean}  &  73.3  &  81.8  &  86.1  &  97.0  &  88.0  &  84.2  &  98.5  &  60.8  & 83.7 \\
{Task Arithmetic}~\cite{ilharco2023editing} & 74.1 & 82.1 & 86.7 & 93.8 & 87.9 & 86.8 & 98.9 & 65.6 & 84.5\\

{Ties-Merging}~\cite{yadav2024ties} &  76.5  &  85.0  &  89.3  &  95.7  &  90.3  &  83.3  &  99.0  &  68.8  & 86.0   \\
PCB-Merging~\cite{du2024parameter} & 75.8 & 86.0 & 88.6 & 96.0 & 88.0 & 90.0 & 99.1 & 70.0 & 86.9 \\

AdaMerging++~\cite{yang2024adamerging}  &79.4 & \textbf{90.3} &91.6 &97.4 &93.4 & \textbf{97.5} & 99.0 &79.2 & 91.0 \\
\midrule
\texttt{SE-Merging} (\textbf{Ours})
& \textbf{81.05}  & 89.27  & \textbf{94.13}  & \textbf{98.11}  & \textbf{94.3}  & 95.64 & \textbf{99.6}  & \textbf{80.48} & \textbf{91.57} \\
\bottomrule
\end{tabular}
}
\end{table*}

\textbf{Performance.} 
To evaluate \texttt{SE-Merging}, we conduct experiments in both the vision and language tasks. 
The experimental settings are detailed in \cref{sec:preliminary}. 
We compare \texttt{SE-Merging} against several baseline methods. 
(1) For vision tasks, we compare \texttt{SE-Merging} with fine-tuned individual models, multi-task learning, and various advanced model merging methods, e.g., Weight Averaging~\cite{wortsman2022modelsoups}, Fisher-Merging~\cite{matena2022fishermerging}, RegMean~\cite{jin2023regmean}, Task Arithmetic~\cite{ilharco2023editing}, TIES-Merging~\cite{yadav2024ties}, PCB-Merging~\cite{du2024parameter}, and AdaMerging~\cite{yang2024adamerging}. 
We do not compare the performance of training-based dynamic model merging methods, e.g., EMR-Merging~\cite{huang2024emrmerging}, Twin-Merging~\cite{lu2024twinmerging}, Representation Surgery~\cite{yang2024representation} and Weight-Ensembling MoE~\cite{tang2024merging}, since these methods require additional training on the test dataset or using a perfect router.
(2) For language tasks, we compare \texttt{SE-Merging} with ine-tuned individual models and several model merging methods, including Weight Averaging~\cite{wortsman2022modelsoups}, Fisher-Merging~\cite{matena2022fishermerging}, RegMean~\cite{jin2023regmean}, Task Arithmetic~\cite{ilharco2023editing} and TIES-Merging~\cite{yadav2024ties}.

For vision tasks, in \cref{tab:performance_vitb32} and \cref{tab:performance_vitl14}, \texttt{SE-Merging} achieves state-of-the-art (SOTA) performance among training-free static methods and even outperforms the training-based static approach, i.e., AdaMerging~\cite{yang2024adamerging}. Specifically, \texttt{SE-Merging} achieves an average improvement of $3.86\%$ on ViT-B/32 and $0.57\%$ on ViT-L/14 compared to AdaMerging across all tasks.
For language tasks, in \cref{tab:performance_gpt2}, \texttt{SE-Merging} achieves a significant performance gain with $6.86\%$ compared to TIES-Merging~\cite{yadav2024ties}. 
These results highlight the effectiveness of \texttt{SE-Merging} in maintaining task-specific expertise and enhancing merging performance without additional training.

\begin{figure*}[!tb]
  \begin{center}
    \includegraphics[width=1.0\textwidth]{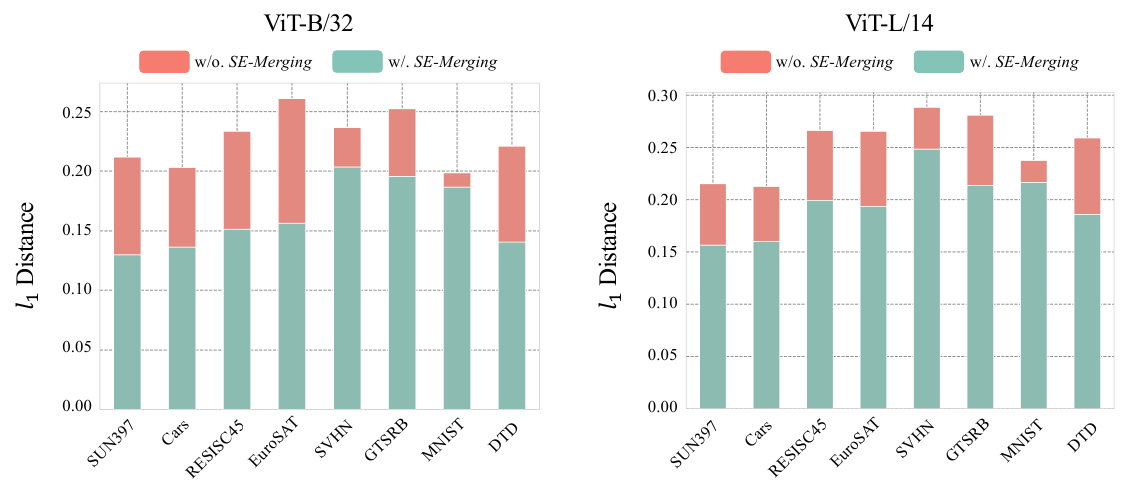}
    \vspace{-1.0em}
    \caption{
    Visualization of the $\ell_1$ distance (``representation bias'' in ~\cref{def:representation_bias}) between the representations of merged model and those of the corresponding fine-tuned model on both ViT-B/32 and ViT-L/14 over samples across all tasks. 
    The \textcolor{green}{green} bars show the decrease of the representation bias after applying the \texttt{SE-Merging} on vanilla Task Arithmetic~\cite{ilharco2023editing}. 
    This indicates that representation of merged model rescaled by \texttt{SE-Merging} is more closer to that of the corresponding fine-tuned model, thereby alleviating the bias problem.
    }
     \label{fig:fig_representation_bias}
  \end{center}
  \vspace{-0.5em}
\end{figure*}

\begin{table*}[htbp]
\centering
\caption{Multi-task performance when merging GPT-2 models on seven text classification tasks.}
\label{tab:performance_gpt2} 
\resizebox{0.93\textwidth}{!}{
\begin{tabular}{l|ccccccc|c}
\toprule
Method         & CoLA & MNLI & MRPC & QNLI & QQP  & RTE  & SST-2 & \textbf{Avg.} \\
    \midrule

    Fine-tuned                                             & 76.8 & 82.1 & 80.4 & 88.3 & 89.6 & 65.3 & 91.2  & 82.0 \\
    \midrule
    Weight Averaging~\cite{wortsman2022modelsoups}   & 55.0 & 55.1 & 51.0 & 57.6 & 76.7 & 44.8 & 52.5  & 56.1 \\
    Fisher Merging~\cite{matena2022fishermerging} & 54.8 & 58.0 & 39.5 & 63.3 & 81.5 & 49.1 & 64.7  & 58.7 \\
    RegMean~\cite{jin2023regmean}           & 61.7 & 70.4 & 65.4 & 69.7 & 78.8 & 56.0 & 79.7  & 68.8 \\
    Task Arithmetic~\cite{ilharco2023editing}        & 68.7 & 68.6 & 69.6 & 70.5 & 81.8 & 47.3 & 83.6  & 70.0 \\
    Ties-Merging~\cite{yadav2024ties}      & 68.4 & 71.4 & 68.4 & 69.6 & 82.4 & 47.7 & 81.8  & 70.0 \\
    \midrule
    \texttt{SE-Merging} (\textbf{Ours})
    & \textbf{72.68} & \textbf{76.14} & \textbf{73.04}  & \textbf{79.17}  & \textbf{85.78}  & \textbf{60.29} & \textbf{90.94} & \textbf{76.86} \\
\bottomrule
\end{tabular}
}
\end{table*}

\textbf{More analysis: \texttt{SE-Merging} alleviates the representation bias problem.} 
Yang et al.~\cite{yang2024representation} introduce the concept of \textit{representation bias}, providing a new perspective to investigate model merging. 
First, we recall the definition of representation bias in ~\cref{def:representation_bias}.

\begin{definition}[Representation Bias~\cite{yang2024representation}]\label{def:representation_bias}
    Suppose there are $T$ tasks, given task $T_i$ with test dataset $\mathcal{D}_i = \{(\boldsymbol{x}_k^i, y_k^i)\}_{k=1}^{N_i}$, the representation bias is defined as the average $\ell_1$ distances between the representations of the merged model $\boldsymbol{\theta}_{\rm Merged}$ and those of the corresponding fine-tuned model at the final output layer $L$ over samples of task $T_i$.
    \begin{equation}
        d = \frac{1}{N_i} \sum_{k=1}^{N_i} \Vert f^{(L)}(\boldsymbol{x}^{i}_k; \boldsymbol{\theta}_{\rm Merged}) - f^{(L)}(\boldsymbol{x}^{i}_k; \boldsymbol{\theta}_{\rm PT} + \lambda\boldsymbol{\tau}_i) \Vert_1
    \end{equation}
    where $\boldsymbol{\theta}_{\rm{Merged}} = \boldsymbol{\theta}_{\rm{PT}} + \lambda \sum_{i}^{T} \boldsymbol{\tau}_i$. 
\end{definition}

As shown in \cref{fig:fig_representation_bias}, \texttt{SE-Merging} significantly alleviates the representation bias problem presented in vanilla Task Arithmetic~\cite{ilharco2023editing}. 
Evidently, when processing samples from a task, \texttt{SE-Merging} aligns the representations of the merged model more closely to those of the corresponding fine-tuned model. 
Through the lens of representation bias, we further demonstrate the effectiveness of \texttt{SE-Merging}. 

\section{Discussions} \label{sec:discussion} 
In this section, we discuss some key questions concerning the empirical findings and the proposed method.
(a) \textbf{The necessity to perform rescaling.}
Since the hypothesis has already identified the task to which the sample belongs, why not leverage the single fine-tuned expert model to perform inference?
The key reason for rescaling and re-merging is that our hypothesis does not hold for all samples. When the hypothesis fails, we can still utilize the expert knowledge within the rescaled merged model to infer correctly.
(b) \textbf{Factors affecting the performance of the method.}
Our method highly depends on the hypothesis, and several key factors affect its performance: (1) Test samples. Although our evaluation is conducted on the full test set across different tasks, evaluating the proposed method on test samples that satisfy the hypothesis and are correctly predicted by the corresponding fine-tuned model will improve performance. (2) Layers of representation. A proper choice of layers better assists in distinguishing the task to which the test sample belongs and enhances the performance. (3) Weight disentanglement. If the fine-tuned models have more disentangled weights, the hypothesis and empirical findings become more evident, leading to improved performance.

\section{Conclusion and Limitations}

We adopted a representation perspective to deepen our understanding of model merging by proposing the Representation Auto-Adaptation Hypothesis (see Hypothesis~\ref{hypo:hypo1}).
Our analysis reveals that model merging achieves multi-task abilities by i) distinguishing samples from different tasks and ii) adapting to the corresponding expert model for each sample.
Building on these insights, we propose \texttt{SE-Merging}, a training-free, sample-wise adaptive framework that dynamically computes scaling coefficients for each sample during inference to enhance the expertise for the corresponding task in the merged model. 
Extensive experiments on vision and language tasks demonstrate that \texttt{SE-Merging} achieves significant performance gains while maintaining compatibility with existing model merging techniques. 
This work not only advances the theoretical understanding of model merging but also provides a practical and efficient solution for multi-task learning in large-scale models.

\textbf{Limitations.}
We note that our theoretical analysis of the Representation Auto-Adaptation Hypothesis relies heavily on the Weight Disentanglement~\cite{ortiz2024task}, which has not yet been rigorously justified in the context of deep neural networks. Providing a theoretical foundation for Weight Disentanglement remains an open challenge, which we will explore in the future. Some existing works address the merging of heterogeneous models~\citep{singh2020otfusion, mergenas, nguyen2023cl_otfusion, zhang2025mah_merge, xu2024tfh_merge}, often utilizing Optimal Transport (OT) techniques~\citep{villani2008ot_book, khamis2024ot_survey, shi2024double_ot, shi2024otclip, shi2023reot}.  However, our paper does not consider the scenario of heterogeneous model merging, leaving it as a future work. We also note that while we have conducted extensive experiments across a wide range of vision and language tasks, further evaluations on broader task categories, such as generative tasks, are currently absent. Such evaluations would help to generalize the applicability.

\section{Acknowledgment}
This work was partially supported by the Shanghai Municipal Science and Technology Major Project (2021SHZDZX0102), Science and Technology Innovation Key R\&D Program of Chongqing (CSTB2024TIAD-STX0035) and  sponsored by MetaLight HK Limited, Hong Kong SAR, China.

\newpage

{
\small
\bibliography{references}
\bibliographystyle{plainnat}
}


\end{document}